\documentclass[sigconf]{acmart}
\usepackage{etoolbox}

\AtBeginEnvironment{tabular}{\Huge}

\renewcommand\footnotetextcopyrightpermission[1]{} 
\usepackage{booktabs} 
\usepackage{multicol}
\usepackage{amsfonts}
\usepackage{adjustbox}
\usepackage{algorithm} 
\usepackage{tabularx}
\usepackage{hyperref}
\usepackage{multirow}
\usepackage{algpseudocode}  
\usepackage{amsmath}  
\AtBeginDocument{%
  \providecommand\BibTeX{{%
    \normalfont B\kern-0.5em{\scshape i\kern-0.25em b}\kern-0.8em\TeX}}}

\begin{document}

\title{Lesion-Inspired Denoising Network: Connecting Medical Image Denoising and Lesion Detection}

\author{Kecheng Chen}
\authornote{Both authors contributed equally to the paper.}
\orcid{0000-0001-6657-3221}

\affiliation{%
  \institution{University of Electronic Science and Technology of China}
  \streetaddress{P.O. Box 1212}
  \city{Chengdu}
  \state{Sichuan}
  \country{China}
  \postcode{43017-6221}
}

\author{Kun Long}
\authornotemark[1]
\affiliation{%
  \institution{University of Electronic Science and Technology of China}
  \city{Chengdu}
  \state{Sichuan}
  \country{China}
}

\author{Yazhou Ren}
\authornote{Co-corresponding authors.}
\affiliation{%
  \institution{University of Electronic Science and Technology of China}
  \city{Chengdu}
  \state{Sichuan}
  \country{China}}
\email{yazhou.ren@uestc.edu.cn}

\author{Jiayu Sun}
\affiliation{%
 \institution{Sichuan University}
 \streetaddress{Rono-Hills}
 \city{Chengdu}
 \state{Sichuan}
 \country{China}}

\author{Xiaorong Pu}
\authornotemark[2]
\affiliation{%
  \institution{University of Electronic Science and Technology of China}
  \streetaddress{30 Shuangqing Rd}
  \city{Chengdu}
  \state{Sichuan}
  \country{China}
  }




\begin{abstract}
Deep learning has achieved notable performance in the denoising task of low-quality medical images and the detection task of lesions, respectively. However, existing low-quality medical image denoising approaches are disconnected from the detection task of lesions. Intuitively, the quality of denoised images will influence the lesion detection accuracy that in turn can be used to affect the denoising performance. To this end, we propose a play-and-plug medical image denoising framework, namely Lesion-Inspired Denoising Network (LIDnet), to collaboratively improve both denoising performance and detection accuracy of denoised medical images. Specifically, we propose to insert the feedback of downstream detection task into existing denoising framework by jointly learning a multi-loss objective.  Instead of using perceptual loss calculated on the entire feature map, a novel region-of-interest (ROI) perceptual loss induced by the lesion detection task is proposed to further connect these two tasks.  To achieve better optimization for overall framework, we propose a customized collaborative training strategy for LIDnet. On consideration of clinical usability and imaging characteristics, three low-dose CT images datasets are used to evaluate  the effectiveness of the proposed LIDnet. Experiments show that, by equipping with LIDnet, both of the denoising and lesion detection performance of baseline methods can be significantly improved.

\end{abstract}




\maketitle

\section{Introduction}
The quality of medical images is crucial for the accurate diagnosis by the physicians \cite{cavaro2010diagnostic,ahmed2011improving}. For medical artificial intelligence (AI) community, quite a few high-level medical image tasks (such as concerned lesion detection \cite{yan2018deeplesion,orlando2018ensemble}, anatomical segmentation \cite{jafari2016skin,sikkandar2020deep}, and multi-modal image registration\cite{arar2020unsupervised,islam2021deep}) rely heavily on extremely high-quality input images, because 1) slight noise perturbation in low-quality images may take unexpected model degradation \cite{hashemi2021cnn}, and 2) some small lesions (e.g., the minute pulmonary nodules) in low-quality images will suffer from severe noise \cite{10.1007/978-3-030-63830-6_36}, leading to the difficulties of the post-processing and the diagnosis. The low-quality medical image will be disturbed by noise and artifacts \cite{prabu2019design}. The researchers therefore focus on low-quality medical image restoration such that the improved images can be used well in potential downstream tasks. 

On consideration of clinical usability and imaging characteristics, the noise removal is the mainstream task for the medical image restoration \cite{sanchez2012medical,jan2005medical}. In various medical images, e.g., computed tomography (CT) image, magnetic resonance imaging (MRI) image, and ultrasonic image, CT image is most sensitive to the noise because the imaging quality will be greatly influenced by the level of radiation dose \cite{polacin1992evaluation,primak2006relationship}. For example, low-dose CT (LDCT) image, which is widely used in the early screening of lung cancer \cite{al2014routine}, will suffer from heavy noise due to the reduction of radiation dose \cite{pearce2012radiation}. It is a challenge for physicians to further analyze more subtle lesion details based on existing low-quality LDCT images.   To this end, low-dose CT denoising task has become the most active medical image denoising orientation recently \cite{chen2017low,chen2017lowzhangyi}.  
General speaking, current medical image denoising approaches represented by LDCT images are roughly categorized into three streams \cite{you2018structurally}. The first stream aims to filter unexpected noise on raw data or sinogram data before image reconstruction \cite{wang2005sinogram,manduca2009projection}. Benefiting from processing raw data directly, classical filtering methods such as bilateral filtering and statistical nonlinear filtering can achieve acceptable performance \cite{balda2012ray}. However, the raw data or sinogram data are usually 
unavailable for researchers \cite{10.1007/978-3-030-63830-6_36}, leading to the limitation. The second stream is the iteration reconstruction-based methods that transform data from sinogram domain to image domain constantly \cite{liu2012adaptive,zhang2016tensor}. They impose some statistical priors in these two domains to minimize an objective function. This stream is usually time-consuming. These two streams can be considered as the pre-processing method.

Thanks to the development of deep learning, the post-processing methods as the third stream achieved better performance. Deep learning-based post-processing methods perform LDCT image denoising in the reconstructed image domain, which has the advantages of the efficiency and the convenience \cite{shan2019competitive,RN400,RN308,RN314,RN429}. However, existing deep learning-based medical image denoising approaches still have some limitations that need to be considered. 

The biggest issue is that existing medical image denoising task is disconnected from downstream tasks. Existing deep learning-based methods typically demonstrate their better denoising performance through multiple evaluation metrics, such as quantitative results (e.g., peak signal-to-noise ratio (PSNR) \cite{RN308}, structure similarity index measure (SSIM) \cite{chen2017low}, texture matching loss (TML) \cite{you2018structurally}) and double-blind scoring experiments \cite{shan2019competitive} (e.g., the scores of noise suppression, contrast, and diagnostic acceptability). However, the actual usability of denoised images for  deep neural network-based downstream medical image tasks (such as concerned lesion detection, anatomical segmentation, and multi-modal image registration) dose not be explored. These downstream tasks are extremely crucial for medicine-related AI communities.

The aforementioned issue may cause that the optimal denoising result is not the optimal one for downstream tasks. For example, existing deep learning-based LDCT images denoising approaches usually calculate the perceptual loss (which can be used to measure the difference of feature space between the denoised result and corresponding ground-truth) on the entire feature map \cite{you2018structurally,shan2019competitive,RN308,RN314,RN429,10.1007/978-3-030-63830-6_36}. However, the lesions detection task, for example, prefers focusing on the local region-of-interest \cite{LONG2021345}, which causes the mismatch of the objective between these two tasks. In this case, the calculation of perceptual loss on local feature maps is a better choice for upstream denoising task.

To tackle these issues, an intuitive idea is to connect the medical image denoising task with downstream tasks. By doing so, the actual usability of denoised images can be reflected explicitly via the output of downstream tasks. The feedback of downstream tasks also can be leveraged by denoising task such that the medical image denoising task can learn a comprehensively optimal pattern regardless of denoising and downstream tasks.  In this paper, 
we concentrate on deep learning-based LDCT image denoising task and downstream deep learning-based lesions detection task, on consideration of clinical use frequency, the availability of datasets and noise characteristics. Following aforementioned idea, we propose a play-and-plug  medical image denoising framework, namely Lesion-Inspired Denoising Network (LIDnet), to collaboratively improve both denoising performance  and the accuracy of detection results for medical images. To be more specific, we propose to connect the medical image denoising task with downstream detection task by inserting the feedback of downstream detection task into existing denoising framework through jointly learning a multi-loss objective. We further connect these two tasks by a novel   region-of-interest (ROI) perceptual loss induced by the lesion detection task, rather than simply using the perceptual loss calculated on the entire feature map.
To achieve better optimization for overall framework,  a customized collaborative training strategy is proposed for LIDnet.

The contributions of this paper have three-folds:
\begin{itemize}
    \item To the best of our knowledge, the  LIDnet framework is the first attempt to connect the medical image denoising task with downstream detection task through jointly learning a multi-loss objective, leading to the collaborative improvements of both denoising effect and detection accuracy 
    
    \item We propose a novel ROI perceptual loss induced by lesions detection task such that the detection results can be further inserted into the denoising task, which better matches the objectives of these two tasks.
    
    \item A customized collaborative training strategy is proposed for LIDnet framework, which can contribute to better optimization orientation for denoising and downstream networks. 
\end{itemize}
\section{Related works}
In this study, the related works includes three aspects, i,e., deep learning-based LDCT image denoising, deep learning-based lesion detection, connection between denoising and downstream tasks.
\subsection{Deep learning-based LDCT image denoising}
LDCT image denoising is the most active orientation in medical image denoising fields, which contributes to its easily accessible datasets \cite{flohr2005image,moen2021low}  and valuable clinical usability. Deep learning-based LDCT image denoising methods have achieved better performance compared with conventional methods. To the best of our knowledge, Chen et. al \cite{chen2017low} firstly adopted a simple convolutional neural network (CNN) to suppress the noise of LDCT images. Sequentially, various CNN-based LDCT image denoising methods were proposed, including the single CNN-based framework \cite{liu2018low}  and generative adversarial network (GAN)-based framework \cite{RN308,RN314,RN429}. Red-CNN \cite{chen2017lowzhangyi} is the typical single CNN baseline denoising method, achieving excellent noise suppression by mean square error loss. Yang et. al \cite{RN276} leveraged Wasserstein-GAN \cite{arjovsky2017wasserstein} (WGAN) and perceptual loss \cite{johnson2016perceptual} to achieve well data style transfer from the LDCT image to the NDCT image.
    Based on this, many researches aimed to improve the ability of the generator of GAN. For example, CPCE \cite{RN308} adopted a conveying path-based CNN. MAP-NN \cite{RN314} proposed a modularized LDCT image denoising deep neural network. For the GAN-based framework, the perceptual loss is calculated on a pre-trained VGG \cite{sengupta2019going} model. Servel recent works also proposed to replaced the perceptual loss with other losses (e.g., SSIM loss \cite{RN278,RN429} and autoencoder loss \cite{RN400}) as they argued that the pre-trained VGG model (trained through ImageNet \cite{5206848}) has the domain mismatch with medical images. In summary, existing methods do not concentrate on the actual usability of denoised image in downstream tasks, which leads to the disconnection with downstream tasks.  Meanwhile, existing approaches do not explicitly care about the performance of the ROIs.

\subsection{Deep learning-based lesion detection}
Girshick et. al \cite{girshick2014rich} first proposed a deep learning-based object detection model: R-CNN, which significantly inproved the accuracy of detection compared to traditional object detection algorithms. Since then, various detection models based on deep learning have been proposed (e.g., Faster R-CNN \cite{ren2015faster}, YoLo \cite{redmon2016you}, center Net \cite{duan2019centernet}), which also opened up a new direction for the detection of lesions in medical images. Anantharaman et. al \cite{anantharaman2018utilizing} applied Mask R-CNN \cite{he2017mask} to the oral pathology domain and used it to detect and segment  herpes labialis and aphthous ulcer. Tajbakhsh et. al \cite{tajbakhsh2019computer} proposed a CNN based detection network for pulmonary embolism detection, which achieved better performance than traditional pulmonary embolism detection algorithms. Zhu et. al \cite{zhu2018deeplung} proposed a deep 3D dual path network for pulmonary nodule detection and classification. U-Net \cite{ronneberger2015u}, a encoder-decoder network, has a wide range of applications in the field of medical images, which is usually used to automatically segment lesions. All the models mentioned above need a high quality image as input, even slight noise may make the detection performance unacceptable. But in clinical, it is very difficult to obtain high quality image. Therefore, denoising  the input image before detection may be a feasible solution to improve the detection result. 

\subsection{Connection between denoising and downstream tasks}
To the best of our knowledge, for medical images, there is no effective connection between medical image denoising and lesion detection methods, which is a crucial motivation for this paper. Liu et al. \cite{liu2020connecting} proposed to connect the denoising task with semantic segmentation for natural images. However, this connection scheme is not customized for medical images because the concern ROIs of medical image can not be considered into this framework. They also do not concentrate on the collaborative optimization for two tasks. 

\section{The proposed Methodology}
\textbf{Preliminary:} For existing LDCT image denoising model and lesions detection model, the training procedures of two models are usually  disconnected. In this paper, we aim to connect the denoising and detection tasks. We denote the training samples on a joint space $X \times Y \times Z$ as $D = \{(x_{i},y_{i},(z_{y\_r}^{i},z_{y\_c}^{i},z_{y\_w}^{i},z_{y\_h}^{i},L_{y}^{i}))\}_{i=1}^{N}$, where $x_{i}$ and $y_{i}$ denote the $i$th LDCT image and corresponding normal-dose CT (NDCT) image, $(z_{y\_r}^{i},z_{y\_c}^{i},z_{y\_w}^{i},z_{y\_h}^{i})$ denote 4 coordinates of bounding box of detection target for the $i$th NDCT image $y_{i}$, and $L_{y}^{i}$ is the classification label of detection target.

For denoising purpose, our goal is to learn a mapping network $F$ that can map $x \rightarrow y$ between two domains $\mathcal{X}$ and $\mathcal{Y}$ with paired training samples $(x,y) \in (X,Y)$. For lesions detection, the process can be roughly divided into two branches, including the generating of region proposal and lesion recognition.

\subsection{Region-Of-Interest Perceptual Loss} \label{3.1}

Before introducing our proposed region-and-interest (ROI) perceptual loss, we review how existing perceptual loss for LDCT denoising is. As pointed by \cite{you2018structurally}, the perceptual loss is very effective for noise removal, leading to the image style of denoised LDCT images very similar to that of NDCT images. Specifically, existing widely-adopted perceptual loss represents the difference of feature space between the denoised LDCT image and corresponding normal-dose one, which is formulated as \cite{you2018structurally}
\begin{equation}
    L_{perceptual\_loss}=L_{pl} = \mathbb{E}_{(x,y)}[\frac{\|\phi_{VGG}(F(x))-\phi_{VGG}(y)\|_{F}^{2}}{whd}],
\end{equation}
where $\phi_{VGG}$ denotes a feature extractor that can generate the output through the 16th convolutional layer of VGG network \cite{sengupta2019going}. $\| \cdot \|_{F}$ denotes the Frobenius norm. $w,h,$ and $d$ denote the width, height, and the number of feature maps, respectively. From Eq. (1), we can observe that the perceptual loss is calculated on  the entire feature map, which we call \textit{global perceptual loss}. Intuitively, this global perceptual loss has a potential limitation: In medical image domain, we usually focus more on local features in the ROI rather than global features, because the concerned lesions (such as the pulmonary nodule and pulmonary embolism) are typically located in local regions of the organ \cite{LONG2021345}. However, existing LDCT image denoising methods tend to achieve an optimal perceptual loss of global features rather than that of local features in the ROI. Thus, the denoising models optimized by the global perceptual loss may not guarantee the optimal denoising results in the ROI.

To tackle the aforementioned limitation, a novel ROI perceptual loss is proposed in this paper. The proposed ROI perceptual loss is a straightforward manner, which not only can be beneficial to the improvement of local features but also can contribute to downstream detection task (that mainly works on the ROI). Interestingly, how can we obtain the ROI in order to compute perceptual loss for local features? Benefiting from the connection of denoising and detection tasks, we propose to leverage the ROIs obtained by the region proposal networks (RPN) \cite{ren2015faster} for the computation of perceptual loss (see Figure 1 for more details). The RPN, a module of detection model, can take an image as the input and outputs a set of region proposals. The process can be formulated as follows
\begin{equation}
    \{(t_{1}, p_{1}),(t_{2},p_{2}),\ldots,(t_{M-1},p_{M-1}),(t_{M},p_{M})\} = RPN(H(F(x))
\end{equation}
where $H(\cdot)$ denotes the feature extraction network (the ResNet50 of Figure 1) in the detection networks. $RPN(\cdot)$ can produce a set of region proposals, consisting of the rectangular bounding box $t_{i}$ composed of 4 coordinate points and the object score $p_{i}$. $M$ denotes the number of region proposals and is usually very large. On consideration of efficiency, we adopt the object score to select top $K$ region proposal for the computation of perceptual loss,
\begin{eqnarray}
    \{(t_{1},p_{1}),(t_{2},p_{2}),\ldots,(t_{K},p_{K})\}= select(\{RPN(H(F(x))_{i=1}^{M}\}), \nonumber  \\s.t. \quad p_{i}>p_{K}.
\end{eqnarray}
This process ensures that the ROIs are obtained relatively meaningful, because the foreground (that is usually concerned objects) has a higher score compared with the background (such as the air regions with black in the CT image). In contrast, existing perceptual loss simply fuses global features and can not explicitly focus on meaningful ROIs. Finally, the ROI perceptual loss on a feature map can be represented as 
\begin{equation}
     L_{ROI\_pl} = \mathbb{E}_{(x,y)}\frac{1}{K}\sum_{i=1}^{K}[\frac{\|T(F(x))_{t_{i}}-T(y)_{t_{i}}\|_{F}^{2}}{whd}],
\end{equation}
where $T(\cdot)$ denotes the feature extractor. For the selection of $T(\cdot)$, existing methods usually adopts the VGG network. In this paper, we propose to adopt the backbone of detection networks as the feature extractor, which is used to calculate the difference between the denoised image and its reference in feature space. 
\begin{figure*}
    \centering
    \includegraphics[width=\textwidth]{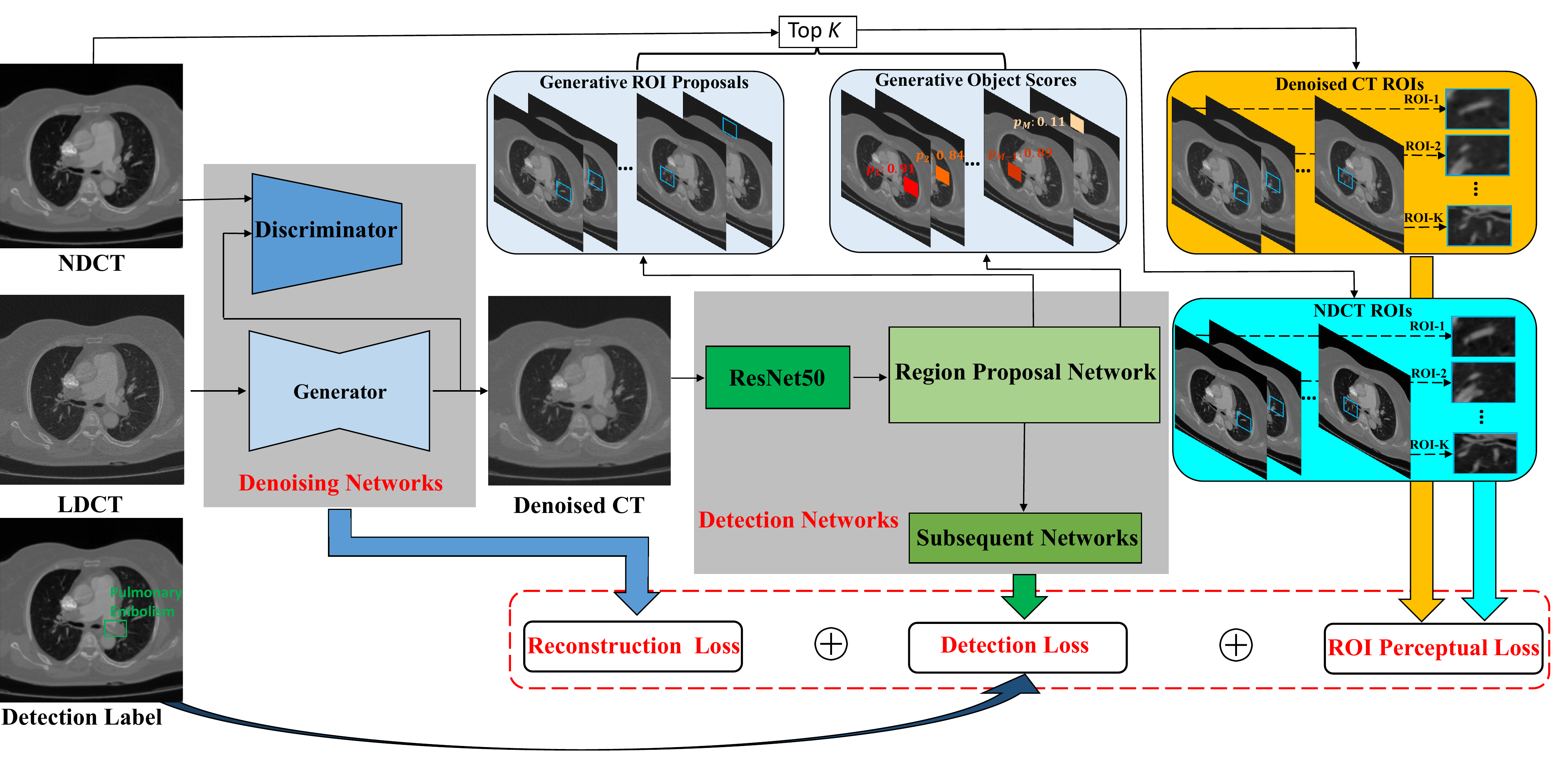}
    \caption{The overall architecture of the proposed LIDnet framework. The denoising networks includes a generator and discriminator. The denoised CT image is produced by the generator. The detection networks includes the backbone network (ResNet50), the region proposal network (RPN), and the subsequent network. The ROI proposals and corresponding object scores are obtained by the RPN. According to the object score, the meaningful foreground ROIs in a denoised CT image and corresponding NDCT are
    selected for the computation of perceptual loss.}
    \label{fig:framework}
\end{figure*}
\subsection{Lesion-Inspired Denoising Network}
Here, we will describe how to connect the denoising networks and lesions detection networks. The motivations of connection have two-fold: First, we mainly leverage the loss of detection networks together with the loss of denoising networks and perceptual loss to jointly optimize the denoising network. This is a simple and effective manner that the denoising network can perceive the feedback of downstream detection task to  collaboratively improve both denoising effect and the accuracy of lesion detection. Second, as described in Section \ref{3.1}, the extracted ROIs are obtained from the detection network for perceptual loss purpose. The ROI perceptual loss is used to optimize the denoising network, which is also an aspect of the connection.

The overall architecture of LIDnet is illustrated in Figure 1. For denoising networks, existing LDCT image denoising methods usually adopt the scheme of generative adversarial network (GAN) \cite{goodfellow2014generative}. GAN is a minimax game with the generator and the discriminator. For the proposed LIDnet, the parameter $\Theta_{G}$ of generator in  denoising networks is optimized by a joint loss as follows:
\begin{eqnarray}\label{total}
    L_{total} &= & \mathbb{E}_{x}[-D(G(x))]  +  \lambda_{1} \mathbb{E}_{(\hat{x},y)}\frac{1}{K}\sum_{i=1}^{K}[\frac{\|T(\hat{x})_{t_{i}}-T(y)_{t_{i}}\|_{F}^{2}}{whd}] \nonumber \\ & + & \lambda_{2}L_{det},
\end{eqnarray}
where the first term is the generator loss, which guarantees
that the data distribution of generator can transform from heavily noisy CT image to corresponding noise-free version. This term is also called \textit{reconstruction loss}, which considers less detailed information. The second and third terms are the proposed ROI perceptual loss and the detection loss respectively. $\lambda_{1}$ and $\lambda_{2}$ are used to balance the two terms.  The third term can be formulated as \cite{ren2015faster}:
\begin{eqnarray}\label{detection_loss}
    L_{det} &=& \mathbb{E}_{(t_{i}^{*},p_{i}^{*},\hat{x})}[Cross\_entropy(p_{i}^{*},S_{1}(H(\hat{x}))) \nonumber \\ &+& L1_{s}(t_{i}^{*},S_{2}(H(\hat{x}))].
\end{eqnarray}
In Eq. (5) and  Eq. (6), $\hat{x}$ denotes the output of denoising networks, $t_{i}^{*}$ and $p_{i}^{*}$ denote the 4 parameterized coordinates of the ground-truth ROI and the label of ground-truth ROI, respectively. The predicted object score $S_{1}(H(\hat{x}))$ and the predicted object coordinates $S_{2}(H(\hat{x}))$ is generated by the subsequent networks of the detection network. $Cross\_
entropy(\cdot)$ denotes the computation of cross entropy loss between predicted object score and corresponding ground-truth label. $L1_{s}$ denotes the smooth L1 loss. More details of detection loss can be found in \cite{ren2015faster}. 

For the loss function of discriminator, we follow the Wasserstein distance with gradient penalty to optimize the discriminator $D$ \cite{gulrajani2017improved}, which can be represented as 
\begin{eqnarray}\label{dis_loss}
    L_{D}  &  = & \mathbb{E}_{\mathbf{\hat{x}}\sim\mathbb{P}_{G}}[D(\hat{x})]-\mathbb{E}_{\mathbf{y} \sim\mathbb{P}_{y}}[D(y)] \nonumber \\
    & + & \lambda \mathbb{E}_{\hat{x}\sim\mathbb{P}_{\hat{x}}} [(\| \nabla_{\hat{x}}D_{\theta}(\hat{x})\|^{2}-1)].
\end{eqnarray}
$\mathbb{P}_{y}$ is the distribution among normal-dose CT images. $\mathbb{P}_{G}$ is the distribution among the output of the generator. $\mathbb{P}_{\hat{x}}$ is defined as a distribution sampling uniformly along straight lines between pairs of points sampled from $\mathbb{P}_{y}$ and $\mathbb{P}_{G}$. $\lambda$ controls the importance of gradient penalty.

\begin{algorithm}[!h]  
  \caption{ The training procedure of proposed LIDnet.}  
  \label{algorithm}  
  \begin{algorithmic}[1]  
    \Require \\
      LDCT images $\mathbf{X}=\{\mathbf{x}_{1},\ldots,\mathbf{x}_{N}\}$, corresponding NDCT images $\mathbf{Y}=\{\mathbf{y}_{1},\ldots,\mathbf{y}_{N}\}$, ground-truth object coordinates of NDCT images $\mathbf{T}^{*}=\{\mathbf{t}_{1}^{*},\ldots,\mathbf{t}_{N}^{*}\}$, and ground-truth object labels of NDCT images $\mathbf{P}^{*}=\{\mathbf{p}_{1}^{*},\ldots,\mathbf{p}_{N}^{*}\}$. Initial denoising network's parameters $\Theta$ and detection networks' parameters $\psi$. \\
      Initial steps for pre-trained detection network $T_{1}$
       \\
      Training steps of denoising network in a round $T_{2}$ \\
      Training steps of detection network based on denoise image in a round ${T_{3}}$
    \Ensure  
      Learned parameters: $\Theta^{*}$ and  $\psi^{*}$.
     \State Freeze $\Theta$
    \For{i=0 to $T_{1}$}
     \State Sample a mini-batch $\mathbf{X}_{d}$, $\mathbf{Y}_{d}$, $\mathbf{T}_{d}^{*}$, and $\mathbf{P}_{d}^{*}$ from $\mathbf{X}$, $\mathbf{Y}$, $\mathbf{T}^{*}$, and $\mathbf{P}^{*}$, respectively.
    \State Optimize $\psi$ with (\ref{detection_loss}) w.r.t. $S_{1}$, $S_{2}$ and $RPN$ on $\mathbf{Y}_{d}$, $\mathbf{T}_{d}^{*}$, and $\mathbf{P}_{d}^{*}$.
    \EndFor
    \While {Stopping criterion is not met}
    \State Unfreeze $\Theta$. Freeze $\psi$ .
    \For {i=0 to $T_{2}$}
    \State Sample a mini-batch $\mathbf{X}_{d}$, $\mathbf{Y}_{d}$, $\mathbf{T}_{d}^{*}$, and $\mathbf{P}_{d}^{*}$ from $\mathbf{X}$, $\mathbf{Y}$, $\mathbf{T}^{*}$, and $\mathbf{P}^{*}$, respectively.
    \State Optimize $\Theta$ with (\ref{total}) (\ref{dis_loss}) w.r.t. D and G (or optimize $\Theta$ with (\ref{eq:denoise-cnn}) w.r.t. G) on  $\mathbf{X}_{d}$, $\mathbf{Y}_{d}$, $\mathbf{T}_{d}^{*}$, and $\mathbf{P}_{d}^{*}$.
    \EndFor
    \State Freeze $\Theta$. Unfreeze $\psi$.
    \For {i=0 to $T_{3}$}
    \State Sample a mini-batch $\mathbf{X}_{d}$, $\mathbf{Y}_{d}$, $\mathbf{T}_{d}^{*}$, and $\mathbf{P}_{d}^{*}$ from $\mathbf{X}$, $\mathbf{Y}$, $\mathbf{T}^{*}$, and $\mathbf{P}^{*}$, respectively.
    \State Compute denoising output by denoising network: $X_{denoise}\leftarrow G(X_{d}$)
    \State Optimize $\psi$ with (\ref{detection_loss}) w.r.t. $S_{1}$, $S_{2}$ and $RPN$ on $\mathbf{X}_{denoise}$, $\mathbf{T}_{d}^{*}$, and $\mathbf{P}_{d}^{*}$.
    \EndFor
    \EndWhile
  \end{algorithmic}  
\end{algorithm}

Except for the scheme of GAN, some state-of-the-art LDCT image denoising methods (such as Red-CNN \cite{chen2017lowzhangyi}) also adopt a single CNN-based network for denoising purpose. They can be also inserted into the proposed LIDnet framework through replacing the GAN-based denoising networks with the single CNN-based network. By doing so, the objective of denoising networks can be formulated as:
\begin{eqnarray}\label{total2}
    L_{total} =  \mathbb{E}_{x,y}[\frac{1}{N}\|F(x)-y\|_{F}^{2}]  +  \lambda_{1} L_{ROI\_pl}  +  \lambda_{2}L_{det},
\label{eq:denoise-cnn}
\end{eqnarray}
where the first term is also the reconstruction loss. $N$ denotes the number of samples in a batch. For the detection networks, the loss function proposed in faster R-CNN is adopted to optimize the parameters. 

From Eq. (\ref{total}) and Eq. (\ref{total2}), the input of detection networks is the denoised CT image produced by upstream denoising networks. This means that if the loss of detection network (the last term) is higher, the denoising quality of denoising network may be worse. To this end, the high detection loss will push the denoising network to learn a better pattern, which can collaboratively improve both denoising performance and the accuracy of detection results.
\subsection{Collaborative Training for LIDnet}
The proposed LIDnet framework involves multiple networks.  We can observe that these networks are mutually constrained. Specifically, the denoising networks need the detection networks to provide ROI proposals in order to calculate the perceptual loss. Meanwhile, the detection networks need the output of the denoising networks as its input. If one of the two networks is not strong enough, the other network will also suffer from terrible optimization direction. To tackle this issue, we proposed a customized collaborative training strategy (CTS) for LIDnet framework, which is similar to the training process of GAN (iterative optimization). To be more specific, we first use high-quality NDCT images to train the detection networks with predefined $T_{1}$ steps, so that the detection networks are optimized in a right direction and could provide acceptable precise ROIs gradually. Then, the training of the denoising networks and detection networks are carried out alternately with multiple rounds. For a round, the denoising networks are trained firstly according to Eq. (\ref{total}) or Eq. (\ref{eq:denoise-cnn}) with predefined $T_{2}$ steps. Here, the ROIs for perceptual loss calculation are provided through the region proposal network of detection networks. After training a certain number of steps, the detection networks are trained again based on the output of denoising networks with predefined $T_{3}$ steps. The overall algorithm is described in Algorithm \ref{algorithm}.

\section{Experiments}
In this section, we carry out experiments to evaluate the effectiveness of the proposed LIDnet framework. To the best of our knowledge, there is no open-source medical image datasets jointly considering CT denoising and detection. To this end, we use NDCT image detection datasets to simulate corresponding LDCT images. The jointly simulated low/normal-dose and detection datasets can be obtained finally. Similar simulated schemes are widely adopted in the fields of LDCT image denoising \cite{chen2017low,gholizadeh2019deep}, because it is unacceptable for patients to be scaned using low and normal-dose radiations twice. The details of used datasets are reported in Table 1.  We adopt a simple and effective LDCT simulation method as shown in  \cite{gholizadeh2019deep}. For PE-CT, L-CT-A and L-CT-G datasets, the noise levels $N_{0}$ are set to 3000, 1000, and 1000, respectively. 
\begin{table}[]
\caption{The details of used datasets.}
\begin{adjustbox}{max width=\columnwidth}
\centering
\begin{tabular}{cccc}
\hline
\textbf{Dataset}         & \textbf{PE-CT} \footnote{https://figshare.com/authors/Mojtaba\_Masoudi/5215238}            &   \textbf{L-CT-A} \footnote{https://wiki.cancerimagingarchive.net/pages/viewpage.action?pageId=70224216}   & \textbf{L-CT-G} \footnote{https://wiki.cancerimagingarchive.net/pages/viewpage.action?pageId=70224216}    \\ \hline \hline
Scaning parts   & Chest            & Lung & Lung \\
Lesion          & Pulmoanry nodule &   Adenocarcinoma   &   Carcinoma   \\
Object Size           & small             &   large   &   large \\
Slice thickness & $\leq$1mm              &   $2mm$   &     $2mm$ \\
Slice interval &   $\leq 1.5mm$           & $0.625mm \sim 5mm$     & $0.625mm \sim 5mm$\\
Resolution &   $512 \times 512$           & $512 \times 512$     & $512 \times 512$\\
Total           & 2304             &   3000   &   3000   \\training           & 2000             &  2500   &   2500
\\ test           & 304             &   500   & 500 \\ \hline   
\end{tabular}
\end{adjustbox}
\end{table}
\begin{table*}[!h]
\centering
\caption{The quantitative results of detection task on three datasets, in terms of AP-50, and AP-75, respectively. For adopted baselines, our proposed framework is used to implement their improved versions. The better score between the baseline and its improved version is bolded with blue. For the AP-50 and AP-75, the higher the better.  }
\begin{adjustbox}{max width=\textwidth}
\begin{tabular}{cccc|cc|cc|cc|cc}
\hline
                                  \textbf{DATASET}                       & \textbf{}                  & WGAN-VGG\cite{RN276} & \textbf{Ours-WGAN-VGG} & CPCE\cite{RN308} & \textbf{Ours-CPCE}      & MAPNN\cite{RN314} & \textbf{Ours-MAPNN} & SSMS \cite{RN278} & \textbf{Ours-SSMS} & Res-HLF\cite{RN429} &  \textbf{Ours-Res-HLF}     \\ \hline \hline
\multicolumn{1}{c||}{\multirow{2}{*}{\textbf{PE-CT}}}                                       & \multicolumn{1}{c||}{AP-50} & 74.56         &  \textcolor{blue}{\textbf{74.59}}            & 74.60         & \textcolor{blue}{\textbf{75.69}} & 75.03          &       \textcolor{blue}{\textbf{76.79}}         &      73.16         &      \textcolor{blue}{\textbf{74.07}}         &    74.34          & \textcolor{blue}{\textbf{76.44}} \\
\multicolumn{1}{c||}{}                                    & \multicolumn{1}{c||}{AP-75} & 25.46         &        \textcolor{blue}{\textbf{26.70}}       & 25.74         &    \textcolor{blue}{\textbf{26.26}} & 25.75          &   \textcolor{blue}{\textbf{27.01}}             &       23.02        &       \textcolor{blue}{\textbf{24.62}}        &         \textcolor{blue}{\textbf{25.74}}         & 24.77 \\ \hline \hline
\multicolumn{1}{c||}{\multirow{2}{*}{\textbf{L-CT-G}}}   & \multicolumn{1}{c||}{AP-50}         &     93.90     &\textcolor{blue}{\textbf{95.31}}               &     91.62      &  \textcolor{blue}{\textbf{93.74}}  &      92.84     & \textcolor{blue}{\textbf{95.14}}               &     91.15          &  \textcolor{blue}{\textbf{93.89}}               &      94.24 & \textcolor{blue}{\textbf{95.28}}   \\
\multicolumn{1}{c||}{}                                    & \multicolumn{1}{c||}{AP-75} &    46.15     & \textcolor{blue}{\textbf{48.14}}         &   35.89            &    \textcolor{blue}{\textbf{46.67}}       &  45.04 & \textcolor{blue}{\textbf{45.53}}          &      38.76         &  \textcolor{blue}{\textbf{44.00}}             &          47.74        & \textcolor{blue}{\textbf{48.90}}          \\ \hline \hline
\multicolumn{1}{c||}{\multirow{2}{*}{\textbf{L-CT-A}}}                                    & \multicolumn{1}{c||}{AP-50} &\textcolor{blue}{\textbf{91.51}}          &91.31          &   88.01       & \textcolor{blue}{\textbf{89.76}}               & 86.98          &\textcolor{blue}{\textbf{91.44}}  & 84.02         &\textcolor{blue}{\textbf{88.50}}           &           89.32       &\textcolor{blue}{\textbf{90.60}}           \\
\multicolumn{1}{c||}{}                                    & \multicolumn{1}{c||}{AP-75} &41.34          &\textcolor{blue}{\textbf{41.48}}          &    41.30      &41.30                & 40.00          &\textcolor{blue}{\textbf{43.21}}  &37.49      & \textcolor{blue}{\textbf{40.39}}              &      39.85            & \textcolor{blue}{\textbf{40.36}}          \\ \hline
\end{tabular}
\end{adjustbox}
\end{table*}

\begin{table*}[!h]\large
\caption{The quantitative results of denoising task on three datasets for the ROIs and the overall image. For adopted baselines, our proposed framework is used to implement their improved versions. The better score between the baseline and its improved version is bolded with blue. For the correlation, homogeneity, RMSR, and energy, the lower the better. For the PSNR and the SSIM,  the higher the better. }
\setlength{\extrarowheight}{6pt}
\begin{adjustbox}{max width=\textwidth}
\begin{tabular}{ccccccc|cccccc|cccccc}
\hline               \multicolumn{1}{c}\textbf{\textbf{DATASET}}
                                   & \multicolumn{6}{c||}{\textbf{PE-CT}}                        & \multicolumn{6}{c||}{\textbf{L-CT-G}}                        & \multicolumn{6}{c}{\textbf{L-CT-A}}                        \\
                                                         \hline \hline
                                   & \multicolumn{3}{c|}{\textbf{ROI}}        & \multicolumn{3}{c||}{\textbf{Overall}} & \multicolumn{3}{c|}{\textbf{ROI}}       & \multicolumn{3}{c||}{\textbf{Overall}} & \multicolumn{3}{c|}{\textbf{ROI}}       & \multicolumn{3}{c}{\textbf{Overall}} \\ \hline
                                   & Correlation & Homogeneity  & \multicolumn{1}{c|}{Energy} & PSNR        & SSIM       & \multicolumn{1}{c||}{RMSE}       & Correlation & Homogeneity & \multicolumn{1}{c|}{Energy} & PSNR        & SSIM       & \multicolumn{1}{c||}{RMSE}       &Correlation & Homogeneity  & \multicolumn{1}{c|}{Energy} & PSNR        & SSIM       & RMSE       \\ \hline \hline
\multicolumn{1}{c|}{WGAN-VGG}      &   0.023   &   \textcolor{blue}{\textbf{0.038}}   &      \textcolor{blue}{\textbf{0.009}}                     &       27.49      &  99.97           &     \multicolumn{1}{c||}{0.046}      &\textcolor{blue}{\textbf{0.048}}      &\textcolor{blue}{\textbf{0.068}}      &0.019                           &26.74             &99.97            & \multicolumn{1}{c||}{0.049}           &\textcolor{blue}{\textbf{0.042}}      &\textcolor{blue}{\textbf{0.085}}      &\textcolor{blue}{\textbf{0.025}}                           &27.18             &99.94             &0.049            \\
\multicolumn{1}{c|}{Ours-WGAN-VGG} &  0.023    &  0.058    &     0.013                      &       \textcolor{blue}{\textbf{35.51}}      &      \textcolor{blue}{\textbf{1.00}}        &     \multicolumn{1}{c||}{ \textcolor{blue}{\textbf{0.017}}}        &0.049      &0.076      &\textcolor{blue}{\textbf{0.018}}                           &\textcolor{blue}{\textbf{30.94}}             &\textcolor{blue}{\textbf{99.99}}            &\multicolumn{1}{c||}{\textcolor{blue}{\textbf{0.032}}}            &0.048      &0.104      &0.029                          &\textcolor{blue}{\textbf{29.13}}             &\textcolor{blue}{\textbf{99.96}}            &\textcolor{blue}{\textbf{0.042}}            \\ \hline \hline
\multicolumn{1}{c|}{CPCE}          &0.021 &0.045 &0.012                       &26.34        &99.97       & \multicolumn{1}{c||}{0.049}        &0.055      &0.090     &0.022                 &28.45            & 99.98           & \multicolumn{1}{c||}{0.042}          &0.049      &0.109      &0.030                           &26.30             &99.93            &0.057            \\
\multicolumn{1}{c|}{Ours-CPCE}     &\textcolor{blue}{\textbf{0.019}} & \textcolor{blue}{\textbf{0.034}} & \textcolor{blue}{\textbf{0.009}}                       &\textcolor{blue}{\textbf{31.76}}        & \textcolor{blue}{\textbf{99.99}}       &    \multicolumn{1}{c||}{\textcolor{blue}{\textbf{0.028}} }   &\textcolor{blue}{\textbf{0.054}}      &\textcolor{blue}{\textbf{0.078}}      &\textcolor{blue}{\textbf{0.018}}       &\textcolor{blue}{\textbf{29.90}}             &\textcolor{blue}{\textbf{99.99}}            & \multicolumn{1}{c||}{\textcolor{blue}{\textbf{0.035}}}            &\textcolor{blue}{\textbf{0.043}}      &\textcolor{blue}{\textbf{0.105}}      &0.030                           &\textcolor{blue}{\textbf{28.01}}             &\textcolor{blue}{\textbf{99.95}}            & \textcolor{blue}{\textbf{0.047}}           \\ \hline \hline
\multicolumn{1}{c|}{MAPNN}         & 0.024 & 0.045 &0.011                       &27.33        &99.97       & \multicolumn{1}{c||}{0.043 }       & 0.052           &0.078      &0.018     &23.37                           &99.91            & \multicolumn{1}{c||}{0.077}            &0.047      &0.102      &0.027                           &28.20             &99.95            &0.046            \\
\multicolumn{1}{c|}{Ours-MAPNN}    & \textcolor{blue}{\textbf{0.018}}  & \textcolor{blue}{\textbf{0.038}} & \textcolor{blue}{\textbf{0.009}}                       & \textcolor{blue}{\textbf{31.85}}       &    \textcolor{blue}{\textbf{99.99}}     & \multicolumn{1}{c||}{\textcolor{blue}{\textbf{0.027}}   }     &\textcolor{blue}{\textbf{0.046}}    &\textcolor{blue}{\textbf{0.077}}      &\textcolor{blue}{\textbf{0.017}}                           &\textcolor{blue}{\textbf{30.70}}             &\textcolor{blue}{\textbf{99.99}}            &\multicolumn{1}{c||}{\textcolor{blue}{\textbf{0.032}}}            &\textcolor{blue}{\textbf{0.045}}      &\textcolor{blue}{\textbf{0.101}}      &0.027                           &\textcolor{blue}{\textbf{28.45}}             &\textcolor{blue}{\textbf{99.96}}            &\textcolor{blue}{\textbf{0.045}}            \\ \hline \hline 
    \multicolumn{1}{c|}{SSMS}          &    0.040  &   0.138   &  0.043                         &     \textcolor{blue}{\textbf{33.89}}        &     99.99      &    \multicolumn{1}{c||}{ \textcolor{blue}{\textbf{0.021}}    }   &0.071      &0.109      & 0.022             &24.55             &99.95            & \multicolumn{1}{c||}{0.061}            &0.047      &0.125      &0.032                           &\textcolor{blue}{\textbf{27.56}}             & 99.94           &\textcolor{blue}{\textbf{0.049}}            \\
\multicolumn{1}{c|}{Ours-SSMS}     &  \textcolor{blue}{\textbf{0.028}}   &               \textcolor{blue}{\textbf{0.087}}                 &    \textcolor{blue}{\textbf{0.021}}         &  31.86  &      99.99   &      \multicolumn{1}{c||}{0.027 }     & \textcolor{blue}{\textbf{0.070 } }    &\textcolor{blue}{\textbf{0.010}}      & 0.022                 &\textcolor{blue}{\textbf{29.84}}   &\textcolor{blue}{\textbf{99.99}}   &\multicolumn{1}{c||}{\textcolor{blue}{\textbf{0.035}}}            &\textcolor{blue}{\textbf{0.046}}      &\textcolor{blue}{\textbf{0.105}}      &\textcolor{blue}{\textbf{0.030}}                           & 25.85            & 99.94            &0.057            \\ \hline \hline
\multicolumn{1}{c|}{Res-HLF}       &  \textcolor{blue}{\textbf{0.023}}    &   0.107   &   0.031                        &     31.35        &    99.99        &      \multicolumn{1}{c||}{0.030  }    &\textcolor{blue}{\textbf{0.025}}      &0.080      &0.022                           &\textcolor{blue}{\textbf{32.30}}             &99.99            & \multicolumn{1}{c||}{0.027}            &\textcolor{blue}{\textbf{0.036}}      &0.105      &0.029                           &29.70             &99.96            &0.049            \\
\multicolumn{1}{c|}{Ours-Res-HLF}  &  0.122    &    \textcolor{blue}{\textbf{0.106}}  &    \textcolor{blue}{\textbf{0.021}}                       &   \textcolor{blue}{\textbf{32.11}}          &    99.99       &      \multicolumn{1}{c||}{\textcolor{blue}{\textbf{0.027}} }  &0.033      &\textcolor{blue}{\textbf{0.063}}      &\textcolor{blue}{\textbf{0.016}}                           &31.67             &\textcolor{blue}{\textbf{99.99}}            & \multicolumn{1}{c||}{\textcolor{blue}{\textbf{0.026}}}            &0.054      & \textcolor{blue}{\textbf{0.097}}     &\textcolor{blue}{\textbf{0.026}}                           &\textcolor{blue}{\textbf{30.78}}             & \textcolor{blue}{\textbf{99.98}}           &\textcolor{blue}{\textbf{0.034}}            \\ \hline 
\end{tabular}
\end{adjustbox}
\end{table*}
\subsection{Baseline Methods and Evaluation Metrics }
We compared our proposed LIDnet framework with following baseline methods, in terms of the quantitative/visual results of denoised images and the detection accuracy of denoised images. The details of baseline methods are described as follows:
\begin{itemize}
     \item \textbf{WGAN-VGG} (2018) \cite{RN276}:  We consider the WGAN-VGG model proposed
in \cite{RN276} as the baseline. WGAN-VGG adopts the Wasserstein-GAN and perceptual loss (computed by the VGG network)
for LDCT image denoising.
    \item \textbf{CPCE} \cite{RN308} (2018): We train a Conveying Path based Convolutional Encoder-decoder (CPCE) network. This network achieves impressive LDCT image denoising performance and also uses the VGG-based perceptual loss for training.
    \item \textbf{MAP-NN} \cite{RN314} (2019): We apply a modularized LDCT image denoising deep neural network (MAP-NN). This network includes multiple equal denoising modules. We choose the version with 3 modules in order to balance the denoising level and the detailed retention. MAP-NN also adopts the VGG-based perceptual loss. \item \textbf{SSMS} \cite{RN278} (2019): This network adopts a Structurally-Sensitive Multi-Scale (SSMS) deep neural network for low-dose CT denoising. Instead of using VGG-based perceptual loss, the structured similarity index measure (SSIM) loss is used to measure the structural and perceptual similarity between two images.
    \item \textbf{Res-HLF} \cite{RN429} (2020): We apply a generative adversarial network with a hybrid loss function for LDCT image denoising. Similar to SSMS, Res-HLF also replaces the VGG-based perceptual loss with  SSIM loss in order to measure the perceptual similarity. In addition, the residual learning is introduced.
\end{itemize}


\begin{figure*}[!ht]
    \centering
    \includegraphics[width=\textwidth]{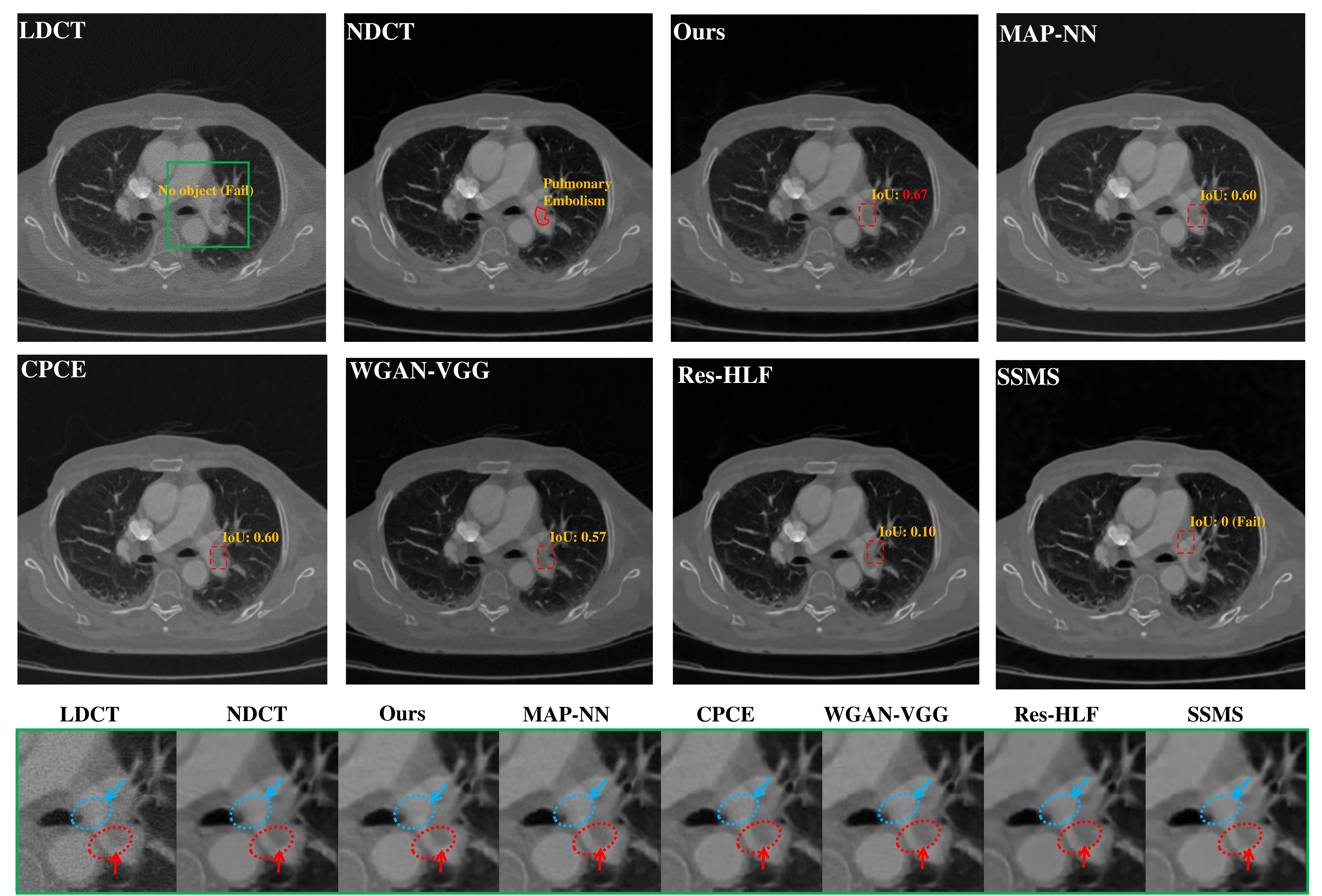}
    \caption{ A visualized pulmonary embolism detection example on PE-CT dataset. The pulmonary embolism lesion is labeled with the red in the NDCT image. For every denoised  CT image, the red dashed box is the bounding box of the detection object. Note that the object can not be detection in the LDCT. For the score of IoU, the higher the better. The green box is a selected observation region in order to analyze the detailed denoising performance. By doing so,  the zoomed-in sub-images for every CT image can be generated and then illustrated in the last row of Figure 2. In the zoomed-in sub-figure, the blue and red dashed circles are two selected regions for better comparison. The two regions are close to the lesion. Please zoom in for better view.  }
    \label{fig:comparision}
\end{figure*}

To evaluate the denoising performance of different methods, we calculate quantitative performance of denoising results on the region-of-interest (ROI) and overall image, respectively. The selected ROIs are from the detection labels (the region of the lesion), which can deliver more useful pathological information. For the ROIs, we use the mean absolute difference of radiomics features between the denoised images and corresponding NDCT images to evaluate the retention capacity of the underlying disease characteristics. The radiomics feature has powerful ability to uncover the microcosmic feature of the lesion in the ROIs. We empirically select three typical radiomics features (including the correlation, homogeneity, and energy) in this paper. The Pyradiomics platform \footnote{https://pyradiomics.readthedocs.io/en/latest/} is used to calculate single values per feature for a region of interest. More detail about computational radiomics features can be found in \cite{van2017computational}. For overall performance, the peak signal-to-noise ratio (PSNR), structured similarity index measure (SSIM), and root mean square error (RMSE) are adopted. 

To evaluate the detection performance of different methods, the Average Precision (AP)-50 and AP-75 are used in this paper. It should be noted that the quantitative detection results for every dataset are calculated through a pre-trained detection network. Specifically, the per-trained detection network is trained based on the NDCT images. Intuitively, if the denoised images is closer than corresponding NDCT image, the score of AP will be higher. 

\subsection{Network Structure}
In this paper, the proposed LIDnet is a play-and-plug framework that aims to collaboratively improve the denoising and detection performances of existing methods. Thus, we respectively adopt 5 baseline methods as the backbone of denoising networks  together with a modified faster RCNN as the detection networks. Note that ResNet50 is empirically used as the backbone of feature extractor. By doing so, 5 LIDnet-based models, namely \textbf{Ours}-\textit{model\_name} (e.g. \textbf{Ours-CPCE}), can be obtained finally. We compare the denoising and detection performances between the baseline methods and their LIDnet-based versions to show the effectiveness of our proposed LIDnet. For all the models, the Adam optimizer is used in a minibatch manner. The batch size is 8. More experimental settings can be found in supplementary materials.
\subsection{Results on the Detection Task}\label{4.3}
5 baseline methods and their LIDnet-based versions are trained on three aforementioned datasets. Note that 5 baseline methods follow their default training protocols. We adopt our proposed collaborative training strategy for LIDnet-based versions. The trained denoising models are used to carry out the denoising task on test set of every dataset. We then impose the pre-trained detection network to evaluate the detection performance of denoised CT images.

The quantitative results of detection task is reported in Table 2. As we can see, compared with all baseline methods, the LIDnet-based versions of baseline methods can achieve better performance in a clear margin, which is reasonable as our proposed LIDnet framework straightforwardly inserts the feedback of detection networks into the upstream denoising networks by jointly learning a multi-loss objective (see Eq. (\ref{total}) and Eq. (\ref{total2})).
We can also observe that in PE-CT dataset (small object), although our proposed LIDnet-based Res-HLF model slightly fall behind its original in term of AP-75, the performance of AP-50 has a very obvious improvement. Meanwhile, the detection of small object
is usually a very difficult task for existing detection models \cite{LONG2021345}, which in turn requires high-quality input images (the denoised images here). We report a visualized pulmonary embolism detection example of PE-CT dataset in Figure 2. The 
lesion location of pulmonary embolism in this CT image is shown in the NDCT image of Figure 2. 
We can notice that most methods (except for SSMS) can improve the detection performance compared with the LDCT image (the object even can not be detected), which is reasonable as the denoised image has less noise perturbation. Compared with all baselines, our proposed model (the LIDnet-based MAP-NN model) achieves better  score of IoU (Intersection over Union, which can evaluate the accuracy of object detection), which results from the connection of denoising networks and detection networks. Interestingly, although we can find very strong denoising effect in the denoised image of the SSMS, the object still can not be detected. We will discuss this phenomenon in Section \ref{4.4}. 
\begin{figure}
    \centering
    \includegraphics[width=\columnwidth]{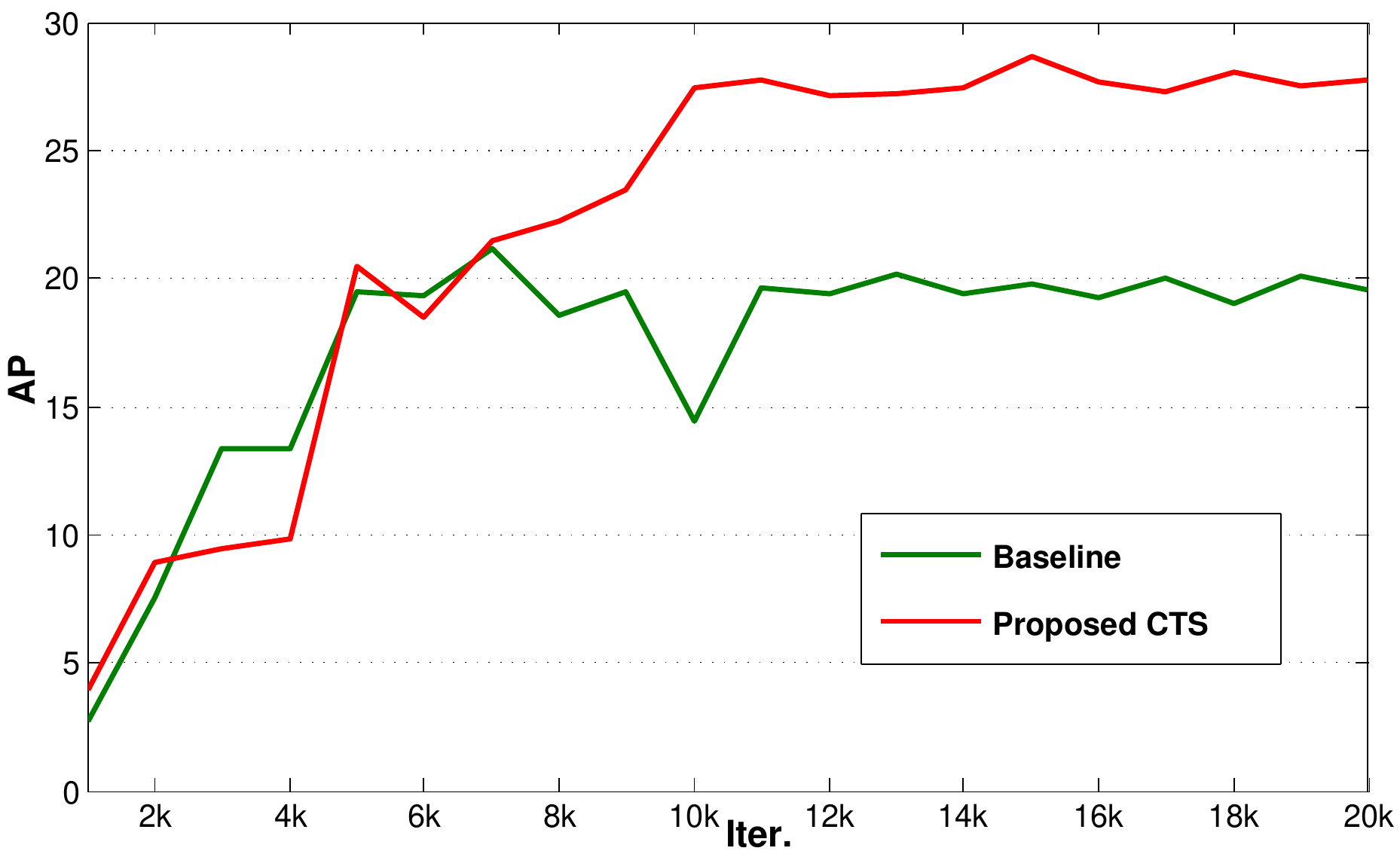}
    \caption{Ablation study on the effectiveness of proposed collaborative training strategy (CTS) for LIDnet. The AP is calculated through the training detection network.}
    \label{fig:CTS2}
\end{figure}
\subsection{Results on the Denoising Task} \label{4.4}
The quantitative results of denoised CT images are reported in Table 3. We have some observations as follows: First, we can observe that the LIDnet-based models have better quantitative radiomics feature of ROIs in most cases, which means that the underlying lesion characteristics are better preserved  compared their baseline methods. On the one hand, the better quantitative performance in the ROIs is beneficial from the proposed ROI perceptual loss, which can make the proposed model concentrate more on the effect of ROIs. On the other hand, detection performance is also influenced by the image quality of the ROIs. 
\begin{figure}[!ht]
    \centering
    \includegraphics[width=\columnwidth]{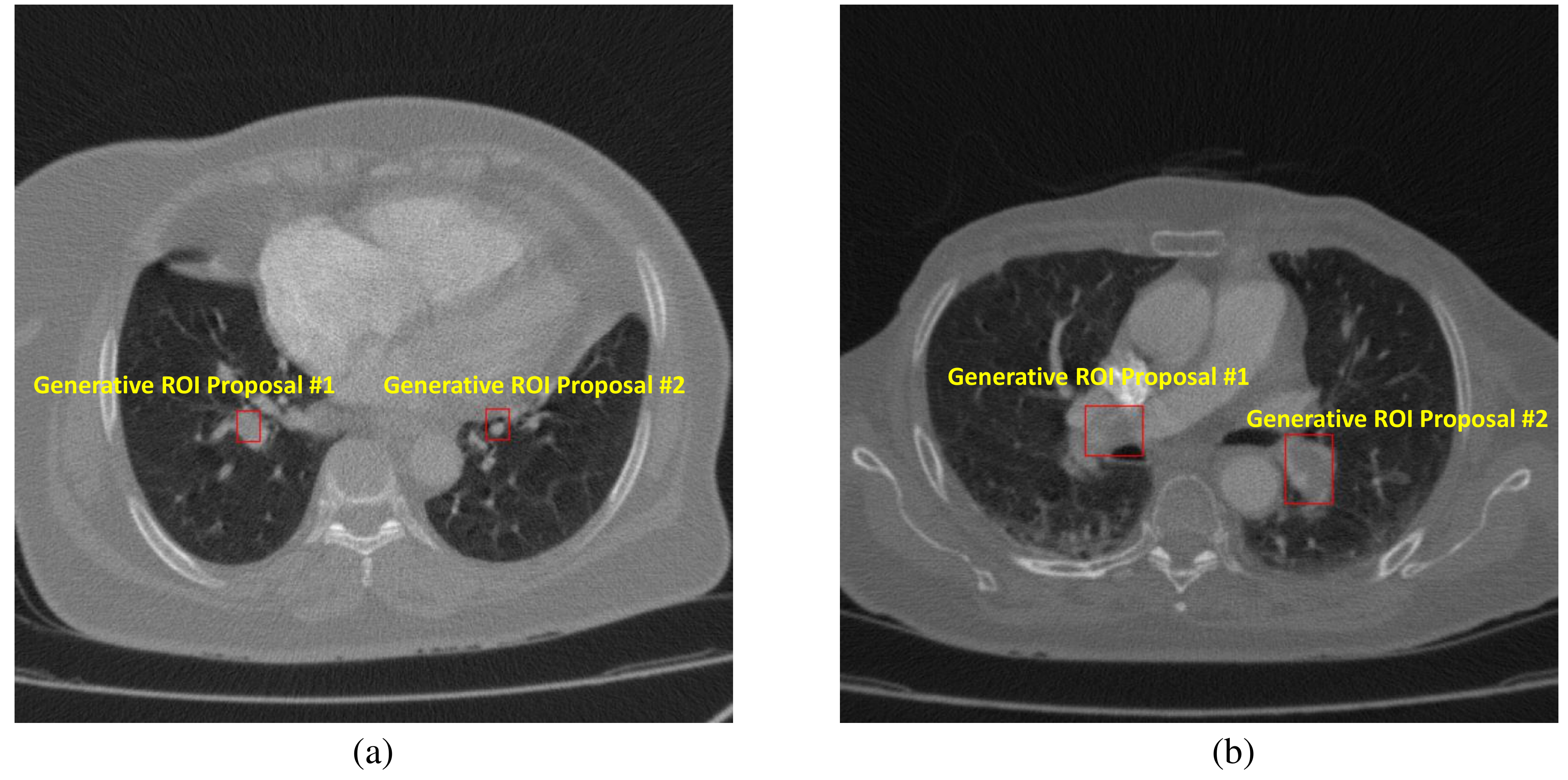}
    \caption{Ablation study on visualized examples for generative ROI proposals. The generative ROI proposals are used for the calculation of ROI perceptual loss. Please zoom in for better view.}
    \label{fig:roi}
\end{figure}
To minimize the detection loss, the proposed model must generate better performance in ROIs. Second,  compared with baseline methods, the LIDnet-based models also improve the overall performances (including PSNR, SSIM, and RMSE) in most cases, which means that the denoised images are closer to the NDCT images in the overall view. Third, we can observe that in the PE-CT dataset, the overall quantitative results of SSMS is better than their LIDnet-based version, especially for PSNR.
However, as mentioned in Section \ref{4.3}, the denoised image of SSMS has extremely worse detection performance. We can observe that the zoomed-in sub-figure of SSMS (see the last row of Figure 2 ) is oversoomth, which usually leads to higher PSNR score. In summary, we conjecture that the oversmooth denoised result dose not contribute to detection task. Interestingly, benefiting from the connection of denoising task and detection task, our proposed models not only improve the detection performance on all datasets but also increase the overall quantitative results in most cases.

The visual zoomed-in results are illustrated in the last row of Figure \ref{fig:comparision}. The lesion can be observed well in this selected region. One has following observations: First, as pointed by red arrow, our proposed method achieves the best artifact suppression compared with other methods, leading to a very clear lesion edge. This may benefit from the
effect of ROI perceptual loss. Second, as shown in blue and red circles, we can find that our proposed method has better noise suppression performance compared with MAP-NN, CPCE and WGAN-VGG. The Res-HLF and SSMS show less noise, but they suffer from the oversmooth problem.

\subsection{Ablation Study}
We are first interested in whether the proposed collaborative training strategy (CTS) is effective. Figure \ref{fig:CTS2} illustrates a comparison between the training process of CTS and that of baseline (which denotes that denoising networks and detection networks are optimized simultaneously). We can observe that  in initial steps, baseline training strategy has higher AP score compared with our proposed CTS. This is reasonable because the denoising network of our proposed CTS dose not be optimized in this period (The denoised image has poor quality). Benefiting from the collaborative training strategy, the denoising and detection networks can be optimized sufficiently, which takes a right direction for training, leading to better performance in the later steps. 

We are also interested in whether the generative ROI proposals (see Section \ref{3.1}) is meaningful for the calculation of ROI perceptual loss. The visualized generative ROI proposals for two input LDCT images are illustrated in Figure \ref{fig:roi}. As shown in Figure 4(a), the generative ROI proposal $\#2$ is indeed located in the meaningful position (nodule). As shown in Figure 4(b), these generative ROI proposals are very similar with the lesions labeled in Figure 2, which is reasonable as the detection network will produce increasingly accuracy ROI proposals with training. Benefiting from the accuracy ROI proposals, the ROI perceptual loss will evaluate the image quality between denoised CT image and corresponding NDCT image in feature space, which can improve the denoising performance in the ROIs (see Table 2). 

\section{Conclusion}
In this paper, we propose a play-and-plug denoising approach for medical images to collaboratively improve both denoising performance and detection accuracy. We insert the feedback of downstream detection task into existing denoising framework by jointly learning a multi-loss objective. A novel ROI perceptual loss is also proposed to further connect these two tasks. The proposed collaborative training strategy is helpful to better optimize the two tasks. On consideration of clinical use frequency, the availability of datasets and noise characteristics, we use three low-dose CT datasets to evaluate the effectiveness of proposed framework.



\bibliographystyle{ACM-Reference-Format}
\bibliography{sample-base}
\appendix
\section{Detail of Architectures and Experimental Settings}
For baseline methods, we tune the hyperparameters in a wide range. For WGAN-VGG, MAP-NN and CPCE, the importance term $\lambda'$ (called in their papers) of the perceptual loss is 100 among three datasets. For SSMS and Res-HLF, the importance term $\beta_{ssim}$ (called in their papers) of SSIM loss is 50 among three datasets. The training was stopped when the model converged.

For the LIDnet-based baseline method, we use the generator of every baseline method as the denoiser of single CNN-based denoising network on considerations of training time and resource consumption. Because we find that, compared with the version of GAN-based network, the single CNN-based network not only can achieve competitive performance but also can significantly reduce the training consumption. The results of a ablation study are reported in Table  4. For the hyperparameters, we choose $\lambda_{1} = 5$  and $\lambda_{2} = 5$ among three datasets. To avoid the oversmooth results, the mean absolute error is used as the reconstruction loss. $K$ is empirically set to 5. For proposed collaborative training strategy, $T_{1} = 4000$, $T_{2}=4000$ and $T_{3}=2000$ .

For the all models, the learning rate of generator is $1 \times 10^{-4}$, and the learning rate of discriminator is $4 \times 10^{-4}$. $\lambda = 10$ for the gradient penalty in Eq. (7). We use two NVIDIA GeForce 1080Ti GPUs to train the all models. The size of minibatch is 8. For the fairness, all models are trained with the entire image as the input, rather than the manner of patches.

For detection networks, the learning rate is set to $5 \times 10^{-3}$. The pre-trained ResNet50 network is used. 
\begin{table}[!h]
\caption{Ablation study about different frameworks of LIDnet-based CPCE model on PE-CT dataset. }
\begin{adjustbox}{max width=\columnwidth}
\begin{tabular}{|c|c|c|c|c|c|c|c|c|c|}
\hline
\textbf{Method} & \multicolumn{2}{c|}{\textbf{Detection}} & \multicolumn{6}{c|}{\textbf{Denoising}}                                                                          & \multirow{2}{*}{}      \\ \cline{1-9}
\textbf{}       & \multicolumn{2}{c|}{\textbf{-}}         & \multicolumn{3}{c|}{\textbf{ROI}}                             & \multicolumn{3}{c|}{\textbf{Overall}}            &                        \\ \hline
\textbf{}       & \textbf{AP-50}     & \textbf{AP-75}     & \textbf{Correlation} & \textbf{Homogeneity} & \textbf{Energy} & \textbf{PSNR}  & \textbf{SSIM}  & \textbf{RMSE}  & \textbf{Training time}   \\ \hline
$Ours-CPCE_{CNN}$   & 75.69              & 26.26              & 0.019                & \textbf{0.034}       & \textbf{0.009}  & \textbf{31.76} & \textbf{99.99} & \textbf{0.028} & \textbf{$\sim$7 hours} \\ \hline
$Ours-CPCE_{GAN}$     & \textbf{75.99}     & \textbf{26.34}     & \textbf{0.016}       & 0.045                & 0.011           & 28.46          & 99.98          & 0.039          & $\sim$32 hours         \\ \hline
\end{tabular}
\end{adjustbox}
\end{table}
\newpage
\section{Additional Visualized Example}
The additional visualized pulmonary embolism detection example is illustrated in Figure 5. As we can see, our  proposed model (the LIDnet-based MAP-NN model) achieves the best detection performance compared with other methods. We further analyze the zoomed-in results in the last row of Figure 5. As shown in red circle, we can find that Res-HLF and SSMS are oversmooth results compared with other models. For the region of the lesion (as circled by blue dashed), our proposed model and Res-HLF achieve the better retention of grayscale level (as pointed by the blue arrow). Instead,
other models relatively weaken the grayscale level, especially for WGAN-VGG. Overall speaking, our proposed model achieves comprehensively better visualized performance compared with other models. 
\begin{figure*}[!ht]
    \centering
    \includegraphics[width=\textwidth]{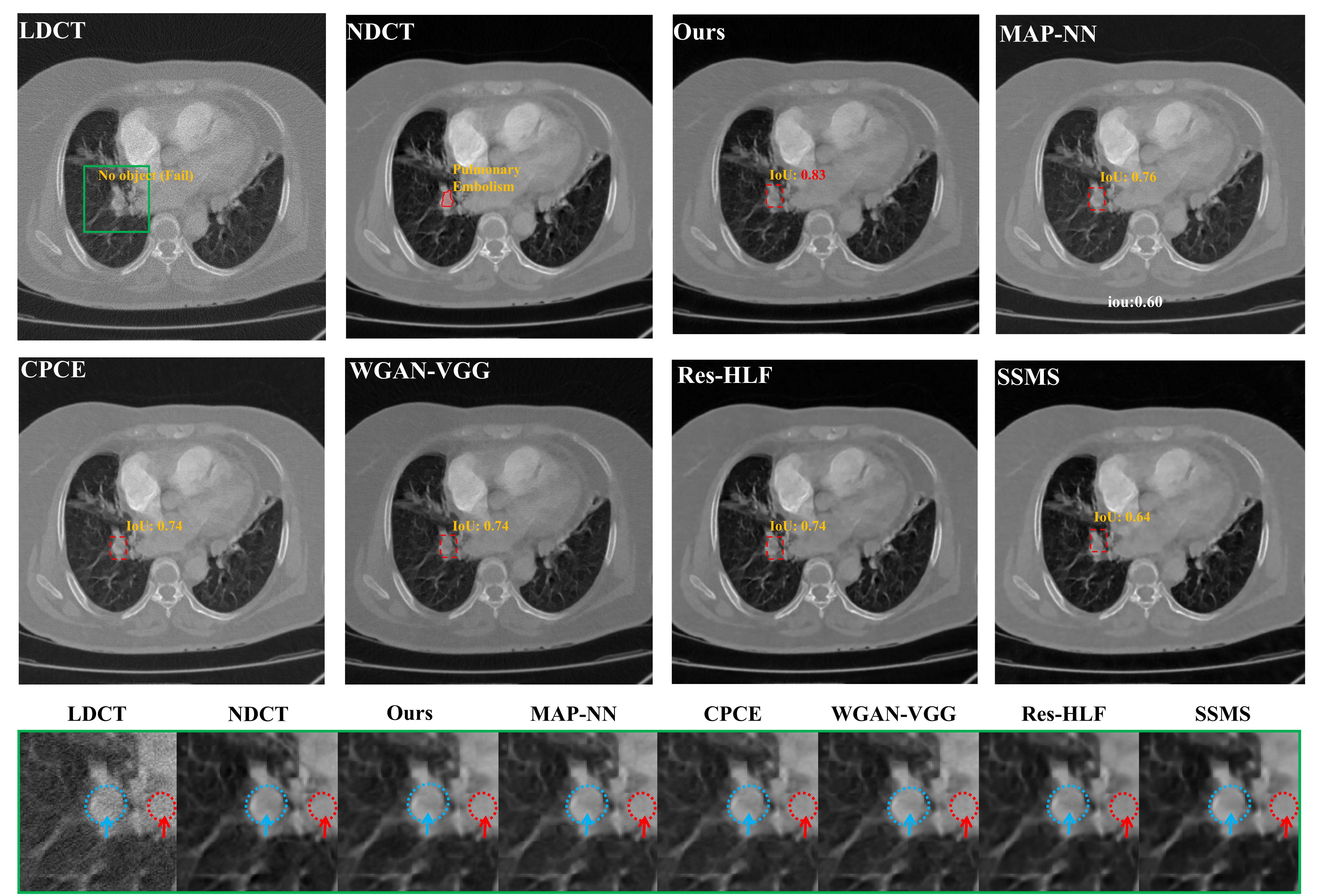}
    \caption{ A visualized pulmonary embolism detection example on PE-CT dataset. The pulmonary embolism lesion is labeled with the red in the NDCT image. For every denoised  CT image, the red dashed box is the bounding box of the detection object. Note that the object can not be detection in the LDCT. For the score of IoU, the higher the better. The green box is a selected observation region in order to analyze the detailed denoising performance. By doing so,  the zoomed-in sub-images for every CT image can be generated and then illustrated in the last row of Figure 5. In the zoomed-in sub-figure, the blue and red dashed circles are two selected regions for better comparison. The two regions are close to the lesion. Please zoom in for better view.  }
    \label{fig:framework}
\end{figure*}
\end{document}